Dona A. Franci[1,2] and Tarik A. Rashid[2]
Department of CS, College of Computer Science and Information Technology, Catholic University in Erbil, Erbil, KR, Iraq
Computer Science and Engineering, University of Kurdistan Hewler, Erbil, KR, Iraq

# CLPB: Chaotic Learner Performance Based Behaviour

## ABSTRACT

This paper presents an enhanced version of the Learner Performance-based Behavior (*LPB*), a novel metaheuristic algorithm inspired by the process of accepting high-school students into various departments at the university. The performance of the *LPB* is not according to the required level. This paper aims to improve the performance of a single objective *LPB* by embedding ten chaotic maps within *LPB* to propose Chaotic LPB (*CLPB*). The proposed algorithm helps in reducing the Processing Time (*PT*), getting closer to the global optima, and bypassing the local optima with the best convergence speed. Another improvement that has been made in *CLPB* is that the best individuals of a sub-population are forced into the interior crossover to improve the quality of solutions. *CLPB* is evaluated against multiple well-known test functions such as classical (*TF1_TF19*) and (*CEC_C06 2019*). Additionally, the results have been compared to the standard *LPB* and several well-known metaheuristic algorithms such as Dragon Fly Algorithm (*DA*), Genetic Algorithm *(GA),* and Particle Swarm Optimization *(PSO)*. Finally, the numerical results show that *CLPB* has been improved with chaotic maps. Furthermore, it is verified that *CLPB* has a great ability to deal with large optimization problems compared to *LPB, GA, DA,* and *PSO*. Overall, Gauss and Tent maps both have a great impact on improving *CLPB*.

## 1. Introduction

The concept of an optimization problem involves searching for the optimal path for solving a particular problem. It can be done by taking into account the objective function of the issue and the various alternatives available. Doing so helps identify the path that will most likely work for the goal [1,2]. Computational Intelligence (CI) is a fast-evolving research area. CI initially was created by Bezdek in the early 1990s [2]. As a term, CI is classified as one of the popular subfields/Branches of AI [2,3].

LPB is a nature-based Metaheuristic algorithm (MA) that is based mainly on the notion of GA. LPB was designed by Rahman and Rashid [2]. It is inspired by the procedure of accepting the graduated students from high school in different departments at the university. However, the main problems encountered in LPB that will be focused on in this paper are: 1) The *PT* of the LPB algorithm is significantly high. 2) The convergence speed and exploitation of the LPB algorithm are not satisfactory. Various mathematical methods are utilized to improve the performance of MAs, such as the chaos theory. This method is a nonlinear system that has dynamic behavior. It has been used by researchers in various areas because of its effectiveness. There exist numerous applications of chaos theory embedded inside MAs for performance enhancement by avoiding trapping into local optima and speeding up the convergence. The use of chaotic maps in MAs is important. However, different phases of the algorithm utilizing different chaotic maps have different effects on metaheuristic algorithms [4]. For instance, earlier chaos theory has been introduced with GA. Since GA has disadvantages such as low convergence speed, so in [5] the authors offered a new GA based on chaotic maps for that Chaotic Genetic Algorithm (CGA) was proposed to address this deficiency. Also, to maximize efficiency, authors in [6] used a genetic algorithm to investigate a single-server inventory system with dynamic customer arrivals and queue-dependent service. This study opens the door for more practical applications of metaheuristics. Moreover, Chaotic maps have been added to the standard Bat Algorithm (BA) to create the Chaotic Bat Algorithm (CBA) [7]. To improve the global search mobility of CBA. Additionally, in [8] again the authors utilized the chaotic theory in the Grey Wolf Optimizer (GWO) algorithm to present the Chaotic Grey Wolf Optimizer (CGWO) algorithm that aims to accelerate its global convergence speed. Also, the authors in [9] proposed a hybridized version of the Whale Optimization Algorithm (WOA) called the chaotic whale optimization algorithm (CWOA) to address the main problem of slow convergence speed. Similarly, chaotic functions have been embedded into another MA, called fitness dependent optimizer (FDO) to create a chaotic fitness dependent optimizer (CFDO) [10]. This work proposes ten chaotic maps inside FDO to improve the performance of the CFDO, with the Tent map being the worst and the Singer map being the best. Additionally, [11] presents a novel edge detection technique for computer vision. The technique combines binary particle swarm optimization (PSO) with an edge-preserving guided filter. Also, in [12] a modified version of the Bird Swarm Algorithm (Dy_DBSA) for blind picture steganalysis is presented. Moreover, The AI-powered method by [13] uses the multi-objective harmony search algorithm. Maximizes the effectiveness of data sanitization keys. Moreover, in

Dona A. Franci[1,2] and Tarik A. Rashid[2]
Department of CS, College of Computer Science and Information Technology, Catholic University in Erbil, Erbil, KR, Iraq
Computer Science and Engineering, University of Kurdistan Hewler, Erbil, KR, Iraq

[14] Brain-computer interface (BCI) systems are being classified using the Harmony search algorithm, who achieve 97% accuracy in the process. This work is close to the work of CFDO in terms of embedding ten chaotic map functions inside the standard algorithm and later on using these chaotic functions to initialize the main population of individuals rather than randomly initializing it in LPB.

In this paper, a new metaheuristic algorithm, chaotic learner performance based-behavior is proposed through the modifications in learner performance based-behavior algorithm. The innovative contributions of this paper to overcome the issues that the LPB has are:

- Improvement in the performance of single objective LPB through modification. Consequently, chaos theory maps are embedded within LPB to propose Chaotic LPB (CLPB), which helps enhance the global convergence speed and quality of solutions and avoid trapping into local optima.
- Reducing the processing time of the algorithm.
- The good individuals of the sub-population are forced to crossover between themselves to improve the quality and diversity of the solutions.
- The importance of CLPB solutions is tested statistically.

The rest sections of the paper are arranged as follows: In *Section 2* the author deliberates on the chaotic maps with their background and characteristics. In *Section 3* the proposed CLPB is described and presented. In *Section 4* the produced experimental results of CLPB are presented. Finally, in *Section 5*, a summary of the work is given.

## 2. Chaotic Maps

In 1972, Edward Lorenz introduced the concept of chaos to the world through his concept of the butterfly effect. This theory aims to make a system more predictable. By understanding this concept, we can keep a system's inputs under control [15]. The concept of chaos theory can be described as that a small change in the initial conditions can lead to a large change later. Most of the available maps are mainly used in the optimization field to make modifications to improve the search efficiency of their algorithms and speed up convergence [7,8,9]. Further details about the chaotic maps and the functions and names of the ten common chaotic maps can be found in [4,7-10].

## 3. Chaotic Learner Performance Based-Behavior CLPB

The Chaotic Learner Performance-based-Behavior algorithm is the modified version of LPB as it has been developed by introducing chaos maps in the LPB algorithm itself. It differs from LPB [2] in two main points. First, within the starting step of the algorithm, the CLPB starts with the creation of the initial population by using chaotic map functions instead of generating the population randomly. This improvement increases the performance of the LPB algorithm as it lacks some vital performance features. The reason for that, is the chaotic maps control the process of choosing the best appropriate random numbers within the range of [0,1] because generating a random number itself cannot always promise to serve in producing optimal solutions as sometimes having a specific random number can be better than other random number and vice versa. Consequently, we need to choose the best random numbers using chaotic map functions that can be valuable for the process of generating the initial population of individuals in CLPB. Second, in CLPB the process of dividing the initial (main) population $M$ into sub-population $O$; the best individuals ($GG$) in $O$ crossover between themselves to improve the quality of solutions. Hence, better individuals (parents) are directed to combine their genes to produce offspring with better results. Therefore, these facts increase the performance of the standard LPB which will lead to the creation of a set of LPB algorithms, or different variants of the chaotic LPB algorithm. The chaotic maps used in CLPB have different mathematical equations. They are divided into ten maps, which are named Chebyshev, Circle, Gauss/mouse, Iterative, Logistic, Piecewise, Sine, Singer, Sinusoidal, and Tent map. Then, CLPB continues the steps of the algorithm similar to the LPB until it gets closer to the promised area. The pseudo-code of the standard LPB algorithm can be found in [2]. Also, the pseudo-code for the modified variant of LPB (CLPB) is shown in Figure 1. The modified parts are coloured in (green and Italics).

### 3.1 The Inspiration of CLPB

Some high school students apply to universities each year to pursue degrees; some are accepted, while others are not. Based on their grade point average (GPA), they are split up into groups based on their GPA. Students transfer to universities after graduation when their GPA is used to assign them to departments. Applications that meet the required range in GPA are accepted by the departments. While some students receive preference, others do not, depending on their GPA. The procedure keeps going until the total number of accepted students meets the department's predetermined limit. Departments and the university may have to make decisions on whether to accept students with lower GPAs or not. The process of grouping the students according to their GPA is done by comparing them

Dona A. Franci[1,2] and Tarik A. Rashid[2]
Department of CS, College of Computer Science and Information Technology, Catholic University in Erbil, Erbil, KR, Iraq
Computer Science and Engineering, University of Kurdistan Hewler, Erbil, KR, Iraq

to the pre-identified GPA of each different department that corresponds to the process of evaluating the fitness (GPA) of individuals in the initial population of CLPB utilizing the fitness function, then, randomly a percentage of the individuals is separated from M called O using a division probability (dp). Thus, this helps in dividing (grouping) the main population into sub-populations (Initial population M, and a separate percentage of the initial population O). Also, the most important features of the CLPB are:
1) The initial population is generated through the use of different chaotic map functions.
2) A percentage of the population is separated.
3) The population is divided into several sub-populations.
4) The good individuals of the population are forced to crossover internally.

## 4. Experimental work of CLPB

The CLPB algorithm has utilized ten common chaotic map functions separately in the LPB interiorly to create the initial population of individuals. This essentially helps in calculating the different positions of individuals by choosing any value that can be assumed as the best random value between 0 and 1; as a result, in this work, the initial point of the chaotic maps was set to (0.7). The different functions of the chaotic maps are proposed by [10]. Each individual in the population of the single objective CLPB algorithm has position and cost characteristics. The fitness for the position of each individual is recognized as cost. The cost of each individual in CLPB is calculated by applying either one of the nineteen classical benchmark functions (TF1-TF19) or one of the ten functions of CEC2019 (CEC01-CEC10) The results are then evaluated against the standard LPB and three popular algorithms in the literature: DA, PSO, and GA. The results for 19 classical benchmark functions for PSO, DA, and GA are taken from [16], and for LPB are taken from [2]. Also, the results for 10 CEC-C06 2019 test functions for all participated algorithms LPB, PSO, and DA are taken [2]. In addition to that, the processing time (PT) of the algorithm for the two groups of test functions is computed to examine how fast is the algorithm in finding the optimal results. Moreover, the significance of the results is proved by using the T.TEST.

```
1. [Initialization]: randomly create a population M
2. [Specify parameters]: specify the number of required
   learners N for a department, crossover rate, and mutation rate
3. [Create Sub-Populations]:
      Use dp parameter to randomly choose a percentage of
      individuals O from M
      Evaluate the fitness of individuals in O
      Depending on their fitness, sort the individuals in O
      (descending order), using one of the sorting methods
      Divide O into two populations, good (individuals with
   high fitness) and bad (individuals with low fitness). Apply
   Crossover between individuals of the good population of
   O, for obtaining better results.
   While the termination condition is not met
      Use dp parameter to randomly choose a percentage
      of individuals O from M
      Evaluate the fitness of individuals in O
      Depending on their fitness, sort the individuals in
      O (descending order), using one of the sorting methods
      Divide O into two populations, good (individuals
      with high fitness) and bad (individuals with low fitness)
      Select half of the individuals of a good population of
      O, using the selection operator
      Apply Crossover between half of the selected individuals
      in the previous step, using the crossover operator
      Find fitness for all individuals in the population M
      Find the highest fitness in good and bad populations
      if an individual from M has fitness <= highest fitness in
         the bad population
            Move it to the bad population (BP)
      else if an individual from M has fitness <= highest fitness
         in a good population
            Move it to the good population (GP)
      else
            Move it to the perfect population (PF)
      end if
   While k <= N
      If PF is not empty
         Select an individual from PF
      else if GP is not empty
         Select an individual from GP
      else
         Select an individual from BP
      end if
      k= k+1;
   end while
4. Crossover
5. Mutation
6. [Termination]: repeat the procedure from step 3 until the
   termination condition is met.
end while
7. [Optional solution]: select the best solution from the perfect
```

Figure 1: CLPB Pseudo Code

### 4.1 Classical benchmark test functions
The benchmark is described as a standard [16,17,18], by which things are measured or evaluated accordingly. In this paper, the CLPB is subjected to the use of two powerful and common sets of benchmarks to evaluate the overall performance enhancement of the CLPB. For more information about the classical benchmarks See [16]. Tables 1 to 3 show the rank results of the comparison between the CLPB algorithm and other popular participating algorithms in the industry GA, PSO, DA, and LPB to demonstrate the differences and optimal results. The parameters for GA, PSO, and DA are given in reference [16] and the parameters for standard LPB are discussed in

Dona A. Franci[1,2] and Tarik A. Rashid[2]
Department of CS, College of Computer Science and Information Technology, Catholic University in Erbil, Erbil, KR, Iraq
Computer Science and Engineering, University of Kurdistan Hewler, Erbil, KR, Iraq

reference [2]. The data for the test functions (TF1-TF19) for LPB, DA, PSO, and GA is from [2,16]. For each TF the results of CLPB1 to CLPB10 have been averaged to one solution to be able to compare it with other contributed algorithms. In Table 1, the overall average performance of the CLPB is 2.47 on a scale of 1 to 5, with 1 being the best and 5 the worst. This ranking takes into consideration that CLPB ranked first three times, second seven times, third four times, fourth two times, and fifth once. See Table 2 for more details. Additionally, the ranking of CLPB by the type of the benchmark function (Unimodal test functions) is 3.1428; (Multi-modal test functions): 2.1667; (Composite test functions): 1.75. Table 3, proves the superiority of the CLPB algorithm against all the participated algorithms in terms of PT in seconds. The PT for the CLPB for optimizing all the functions is much smaller. This is noticeable as the CLPB algorithm has ranked 1st (superior) in 15 out of 17 applicable functions and the difference is significantly large.

### 4.2 CEC-C06 2019 benchmark test function

The second set of benchmark functions is CEC-C06 2019 test functions. The author of with work has examined this benchmark on the CLPB algorithm against other participating algorithms to show the ability of the CLPB algorithm in solving large-scale optimization problems [17]. The CEC-C06 test functions are designed to evaluate the various functions of an algorithm on a horizontal slice of the convergence plot [18]. All the CEC-C06 2019 functions are scalable and all global optimums of these functions were united towards point 1[2]. In [17] the CEC-C06 2019 test function names, dimensions, and ranges can be found.

**Table 1** Ranking of Optimization Algorithms Results (Average fitness) obtained applying CLPB, LPB, DA, PSO, and GA to the classical benchmark functions "TF1-TF19" [23].

| Test Func | 1st | 2nd | 3rd | 4th | 5th | Rank | Subtotal |
|---|---|---|---|---|---|---|---|
| TF1 | DA | PSO | CLPB | LPB | GA | 3 | |
| TF2 | DA | PSO | LPB | CLPB | GA | 4 | |
| TF3 | DA | PSO | CLPB | LPB | GA | 3 | |
| TF4 | DA | PSO | CLPB | LPB | GA | 3 | |
| TF5 | DA | CLPB | LPB | PSO | GA | 2 | |
| TF6 | PSO | DA | CLPB | LPB | GA | 3 | |
| TF7 | LPB | PSO | DA | CLPB | GA | 4 | 22 |
| TF8 | PSO | LPB | GA | DA | CLPB | 5 | |
| TF9 | LPB | CLPB | PSO | DA | GA | 2 | |
| TF10 | LPB | CLPB | DA | PSO | GA | 2 | |
| TF11 | CLPB | LPB | PSO | DA | GA | 1 | |
| TF12 | PSO | CLPB | LPB | DA | GA | 2 | |
| TF13 | CLPB | LPB | DA,PSO | | GA | 1 | 13 |
| TF15 | CLPB | LPB | GA | PSO | DA | 1 | |
| TF16 | LPB | CLPB | PSO | GA | DA | 2 | |
| TF17 | LPB | CLPB | PSO | GA | DA | 2 | |
| TF18 | LPB | CLPB | GA | PSO | DA | 2 | 7 |
| Total: | | | | | | | 42 |
| Overall Rank (Ave. of CLPB rank): | | | | | | | 42/17=2.4705 |
| TF1–TF7(Unimodal TF): | | | | | | | 22/7 = 3.1428 |
| TF8-TF13(Multi-modal TF): | | | | | | | 13/6 = 2.1667 |
| TF15-TF18(Composite TF): | | | | | | | 7/4 = 1.75 |

**Table 2** Details on different ranks of CLPB [23].

| The number of times CLPB has ranked First, Second, Third, Fourth, and Fifth | |
|---|---|
| Rank1 | 3 Times |
| Rank2 | 7 Times |
| Rank3 | 4 times |
| Rank4 | 2 Times |
| Rank5 | 1 Time |

The results of the CEC-C06 2019 test functions of the CLPB algorithm with the participated algorithms (LPB, DA, GA, and PSO) are shown in Table 4. In addition to the CLPB algorithm, the author of the work evaluated LPB, DA, GA, and PSO against the benchmarks, to compare them to the CLPB. For each test function in Table 4, Bold results indicate superior outcomes. The value of the metric standard deviation for the CLPB algorithm in almost all the CEC-C06 2019 test functions is smaller than DA, PSO, GA, and LPB. In contrast, according to the value of the metric, average the LPB showed its superiority in almost all of the functions compared to the participated algorithms. However, CLBP in CEC03 has the same average and standard deviation compared to LPB but with a significant difference in processing time (2.101408) seconds for CLPB while LPB is (144.194876) seconds. In addition to that, CLPB has proved its enormous superiority in the processing time of all the test functions of CEC-C06 2019 in comparison to all the contributed algorithms (LPB, DA, PSO, and GA).

### 4.3 Statistical Tests

Assessing the performance is not accurate enough

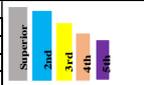

**Table 3** Comparison of PT results of the classical benchmark function between CLPB, LPB, DA, PSO, and GA

| Test Func. | Alg. | PT (Seconds) | Test Func. | Alg. | PT (Seconds) | Test Func. | Alg. | PT (Seconds) | Test Func. | Alg. | PT (Seconds) |
|---|---|---|---|---|---|---|---|---|---|---|---|
| TF1 | CLPB | 2.54764058 | TF6 | CLPB | 2.989005063 | TF11 | CLPB | 3.15101322 | TF16 | CLPB | 2.98629083 |
| | LPB | 160.840946 | | LPB | 157.547318 | | LPB | 130.664299 | | LPB | 181.858429 |
| | DA | 1445.243327 | | DA | 1550.130722 | | DA | 1210.086084 | | DA | 969.827007 |
| | PSO | 249.66503 | | PSO | 2.795879 | | PSO | 9.429028 | | PSO | 4.247415 |
| | GA | 65.422226 | | GA | 51.284046 | | GA | 56.656545 | | GA | 80.998874 |
| TF2 | CLPB | 3.161507773 | TF7 | CLPB | 2.450696973 | TF12 | CLPB | 3.17318701 | TF17 | CLPB | 2.48447188 |
| | LPB | 169.076368 | | LPB | 158.642028 | | LPB | 140.837076 | | LPB | 141.213291 |
| | DA | 1259.496468 | | DA | 1593.877054 | | DA | 1464.060419 | | DA | 1018.757437 |
| | PSO | 3.826913 | | PSO | 8.982616 | | PSO | 22.898798 | | PSO | 2.607163 |
| | GA | 55.040008 | | GA | 56.555067 | | GA | 102.745164 | | GA | 50.990811 |
| TF3 | CLPB | 3.107412447 | TF8 | CLPB | 2.813955769 | TF13 | CLPB | 3.712100023 | TF18 | CLPB | 2.785787777 |
| | LPB | 202.408611 | | LPB | 162.354305 | | LPB | 139.449467 | | LPB | 180.663489 |
| | DA | 1216.762524 | | DA | 1738.794894 | | DA | 1339.438272 | | DA | 1001.716543 |
| | PSO | 12.702411 | | PSO | 8.266467 | | PSO | 16.752814 | | PSO | 2.718852 |
| | GA | 80.126424 | | GA | 55.234252 | | GA | 103.377836 | | GA | 80.273981 |
| TF4 | CLPB | 2.818270893 | TF9 | CLPB | 2.51771657 | TF14 | CLPB | N/A | TF19 | CLPB | N/A |
| | LPB | 191.301934 | | LPB | 159.074029 | | LPB | 170.207352 | | LPB | 169.415055 |
| | DA | 1399.01481 | | DA | 1638.957037 | | DA | 1034.450489 | | DA | 1312.805448 |
| | PSO | 2.877756 | | PSO | 4.816792 | | PSO | 86.298548 | | PSO | 8.952319 |
| | GA | 63.099468 | | GA | 84.833759 | | GA | 152.142368 | | GA | 77.905123 |
| TF5 | CLPB | 3.264821983 | TF10 | CLPB | 3.2636223 | TF15 | CLPB | 3.284513617 | | | |
| | LPB | 130.846636 | | LPB | 148.431567 | | LPB | 247.224271 | | | |
| | DA | 1707.285731 | | DA | 1297.325669 | | DA | 1659.6524 | | | |
| | PSO | 5.224432 | | PSO | 8.013542 | | PSO | 8.250347 | | | |
| | GA | 55.818782 | | GA | 84.666823 | | GA | 54.974533 | | | |

Dona A. Franci[1,2] and Tarik A. Rashid[2]
Department of CS, College of Computer Science and Information Technology, Catholic University in Erbil, Erbil, KR, Iraq
Computer Science and Engineering, University of Kurdistan Hewler, Erbil, KR, Iraq

since it only relies on the standard deviation, average, and PT. To evaluate the CLPB algorithm, the statistical test is used to verify the importance of the results statistically. As a result, almost all the ten chaotic maps could enhance the performance of CLPB. Precisely, Gauss/mouse map and Tent map. This is because these chaotic maps can improve the exploitation and exploration capability of CLPB and speed up the convergence. In the case of [2], it was proved that the results of the LPB are statistically significant compared to DA, PSO, and GA. This means there is no need to compare the CLPB algorithm with DA, PSO, and GA statistically since the CLPB algorithm proved its superiority against the LPB in most of the test functions and with the use of ten chaotic maps.

## 5 Conclusion

In this paper, the author aims to modify the LPB algorithm for better performance. The modification is taking place by introducing 10 chaotic map functions inside LPB to propose chaotic LPB (CLPB), Also, the good individuals of the population are forced to local crossover. Then the proposed algorithm is tested on TF1_TF19 and CEC-C06 2019 benchmarks. The results of these benchmarks are compared with the standard LPB and other participating MAs like PSO, DA, and GA. The use of chaotic maps helped in improving CLPB. The rank results of the classical benchmarks show that the global average performance of CLPB is (2.47) on a scale of 1–5 being 1 the best and 5 being the worst. In terms of the PT, CLPB outperforms the LPB in all cases of both types of test functions. Also, CLPB in almost all cases has superior PT in comparison with the DA, GA, and PSO. Outcomes show that the chaotic maps helped in improving the exploitation phase of LPB since LPB was not up to the required level. CLPB has a brilliant ability in terms of exploration and in avoiding many local optima compared to GA, DA, and PSO while, CLPB showed a slight difference compared to LPB. In general, according to the findings of the CEC-C06 2019 benchmark functions results, CLPB outperforms DA, PSO, GA, and LPB in terms of solving large-scale optimization issues. Moreover, the significance of the results of the proposed work is proved statistically. For future works, Investigation on LPB with advanced algorithms, including Goose [19], Fitness Dependent Optimizer [20], ANA [21], FOX [22], Donkey and Smuggler Optimization, [23]and Child Drawing Development Optimization [24], is at the forefront of future research in computational intelligence and optimization methodologies.

**Table 4 Comparison of results of the CEC-C06 2019 benchmark function between CLPB, LPB, DA, PSO, and GA benchmark functions**

| Test Function | Metrics | DA | CLPB | PSO | GA | LPB |
|---|---|---|---|---|---|---|
| CEC01 | Ave. | 5.43E+08 | 1151565.313 | 1.47E+12 | **5.32E+04** | 7494381363.65768 |
| | Std. | 6.69E+08 | 711637.2073 | 1.32E+12 | **7.04E+04** | 8138223463.28023 |
| | PT in sec. | 2034.95887 | **9.66658189** | 382.330436 | N/A | 377.373846 |
| CEC02 | Ave. | 78.0368 | **17.34929** | 15183.91348 | 17.3502 | 17.63898 |
| | Std. | 87.7888 | **0.00574** | 3729.553229 | 17.3491 | 0.31898 |
| | PT in sec. | 2122.108475 | **2.126793** | 6.064791 | N/A | 140.912536 |
| CEC03 | Ave. | 13.7026 | **12.7024** | 12.70240422 | **12.7024** | **12.7024** |
| | Std. | 0.0007 | **0** | 9.03E-15 | 13.7024 | **0** |
| | PT in sec. | 2223.799974 | **2.101408** | 8.90197 | N/A | 144.194876 |
| CEC04 | Ave. | 344.3561 | 3280.92051 | **16.80077558** | 6.23E+04 | 77.90824 |
| | Std. | 414.0982 | **6.59052** | 8.199076134 | 61986.61 | 29.88519 |
| | PT in sec. | 1720.974833 | **2.391012** | 5.179151 | N/A | 137.305797 |
| CEC05 | Ave. | 2.5572 | 1.53148 | **1.138264955** | 7.5396 | 1.18822 |
| | Std. | 0.3245 | 0.12728 | **0.089389848** | 7.2765 | 0.10945 |
| | PT in sec. | 1722.243949 | **2.525934** | 5.370252 | N/A | 138.406681 |
| CEC06 | Ave. | 9.8955 | 5.03164 | 9.305312443 | 7.4005 | **3.73895** |
| | Std. | 1.6404 | 0.93160 | 1.69E+00 | 6.6877 | **0.82305** |
| | PT in sec. | 1401.682147 | **5.654304** | 131.167162 | N/A | 142.041586 |
| CEC07 | Ave. | 578.9531 | 201.42422 | 160.6863065 | 791.742 | **145.28775** |
| | Std. | 329.3983 | 109.39402 | **104.2035197** | 697.8964 | 177.8949 |
| | PT in sec. | 1376.289834 | **2.368611** | 5.436392 | N/A | 122.135692 |
| CEC08 | Ave. | 6.8734 | 5.26162 | 5.224137165 | 6.1004 | **4.88769** |
| | Std. | 0.5015 | 0.50994 | 0.786760649 | 5.8228 | **0.67942** |
| | PT in sec. | 1802.883649 | **2.52482** | 5.527832 | N/A | 138.20745 |
| CEC09 | Ave. | 6.0467 | 145.28040 | **2.373279266** | 5.31E+03 | 2.89429 |
| | Std. | 2.871 | **0.00075** | 0.018437068 | 5.29E+03 | 0.23138 |
| | PT in sec. | 1365.799778 | 9.404296 | **4.44688** | N/A | 141.699472 |
| CEC10 | Ave. | 21.2604 | 20.09491 | 20.28063455 | 20.1059 | **20.00179** |
| | Std. | 0.1715 | 0.21368 | 0.128530895 | 20.0236 | **0.00233** |
| | PT in sec. | 1699.088096 | **2.54749404** | 9.462923 | N/A | 147.995515 |

**Acknowledgments:**
The authors would like to thank the University of Kurdistan-Hewler for providing facilities for this research work.
**Funding:** This study was not funded.
**Compliance with Ethical Standards Conflict of Interest:** The authors declare that they have no conflict of interest.
**Ethical Approval:** This article does not contain any studies with human participants or animals performed by any of the authors.
**Data Availability Statement:** The datasets generated during and/or analysed during the current study are not publicly available but are

Dona A. Franci[1,2] and Tarik A. Rashid[2]
Department of CS, College of Computer Science and Information Technology, Catholic University in Erbil, Erbil, KR, Iraq
Computer Science and Engineering, University of Kurdistan Hewler, Erbil, KR, Iraq